\documentclass[letterpaper, 10 pt, conference]{ieeeconf}  % Comment this line out if you need a4paper

\IEEEoverridecommandlockouts                              % This command is only needed if 
\usepackage{mathptmx} % assumes new font selection scheme installed
\usepackage{times} % assumes new font selection scheme installed
\usepackage{amsmath} % assumes amsmath package installed
\allowdisplaybreaks
\usepackage{amssymb}  % assumes amsmath package installed
\usepackage{xcolor}

\usepackage{amsfonts}
\usepackage{graphicx}
\usepackage{textcomp}
\usepackage{subcaption}
\usepackage{algorithm}
\allowdisplaybreaks
\usepackage{algpseudocode}
\usepackage{diagbox}
\usepackage[font=footnotesize]{caption}
\usepackage{cite}
\usepackage[hidelinks]{hyperref}

\usepackage{graphicx}  % Use only graphicx, not epsfig
\usepackage{newtxtext,newtxmath}  % Better Times fonts

\title{\LARGE \bf
Automated Generation of Diverse Courses of Actions for Multi-Agent Operations using Binary Optimization and Graph Learning
\thanks{*This work was supported by the Office of Naval Research (ONR) grant N00014-24-1-2003. Any opinions, findings, conclusions, or recommendations expressed in this paper are those of the authors and do not necessarily reflect the views of ONR. }

\thanks{Authors $^{1}, ^{2}, ^{4}$ are with the Department of Mechanical and Aerospace Engineering, University at Buffalo, Buffalo, NY, USA  {\tt\small \{prithvid, ehsanesf, soumacho\}@buffalo.edu}}%

\thanks{Author $^{3}$ is with the Department of Computer Science and Engineering, University at Buffalo, Buffalo, NY, USA {\tt\small kdantu@buffalo.edu}}

\thanks{\textsuperscript{\textdagger} Corresponding author {\tt\small soumacho@buffalo.edu}}

\thanks{Copyright\textcopyright2025 IEEE. Personal use of this material is permitted.
Permission from IEEE must be obtained for all other uses, in any current or
future media, including reprinting/republishing this material for advertising
or promotional purposes, creating new collective works, for resale or
redistribution to servers or lists, or reuse of any copyrighted component
of this work in other works.}
}

\author{Prithvi Poddar$^{1}$, Ehsan Tarkesh  Esfahani$^{2}$, Karthik Dantu$^{3}$ and Souma Chowdhury$^{4,}$\textsuperscript{\textdagger}% <-this % stops a space
}

\begin{document}
\newcommand{\algo}{GenCOA }

\maketitle
\thispagestyle{empty}
\pagestyle{empty}

%%%%%%%%%%%%%%%%%%%%%%%%%%%%%%%%%%%%%%%%%%%%%%%%%%%%%%%%%%%%%%%%%%%%%%%%%%%%%%%%
\begin{abstract}
Operations in disaster response, search \& rescue, and military missions that involve multiple agents demand automated processes to support the planning of the courses of action (COA). Moreover, traverse-affecting changes in the environment (rain, snow, blockades, etc.) may impact the expected performance of a COA, making it desirable to have a pool of COAs that are diverse in task distributions across agents. Further, variations in agent capabilities, which could be human crews and/or autonomous systems, present practical opportunities and computational challenges to the planning process. This paper presents a new theoretical formulation and computational framework to generate such diverse pools of COAs for operations with soft variations in agent-task compatibility. Key to the problem formulation is a graph abstraction of the task space and the pool of COAs itself to quantify its diversity. Formulating the COAs as a centralized multi-robot task allocation problem, a genetic algorithm is used for (order-ignoring) allocations of tasks to each agent that jointly maximize diversity within the COA pool and overall compatibility of the agent-task mappings. A graph neural network is trained using a policy gradient approach to then perform single agent task sequencing in each COA, which maximizes completion rates adaptive to task features. Our tests of the COA generation process in a simulated environment demonstrate significant performance gain over a random walk baseline, small optimality gap in task sequencing, and execution time of about 50 minutes to plan up to 20 COAs for 5 agent/100 task operations.

\end{abstract}

%%%%%%%%%%%%%%%%%%%%%%%%%%%%%%%%%%%%%%%%%%%%%%%%%%%%%%%%%%%%%%%%%%%%%%%%%%%%%%%%
\section{INTRODUCTION}

\begin{figure}[] %!t
    \centering
    \includegraphics[width=\columnwidth]{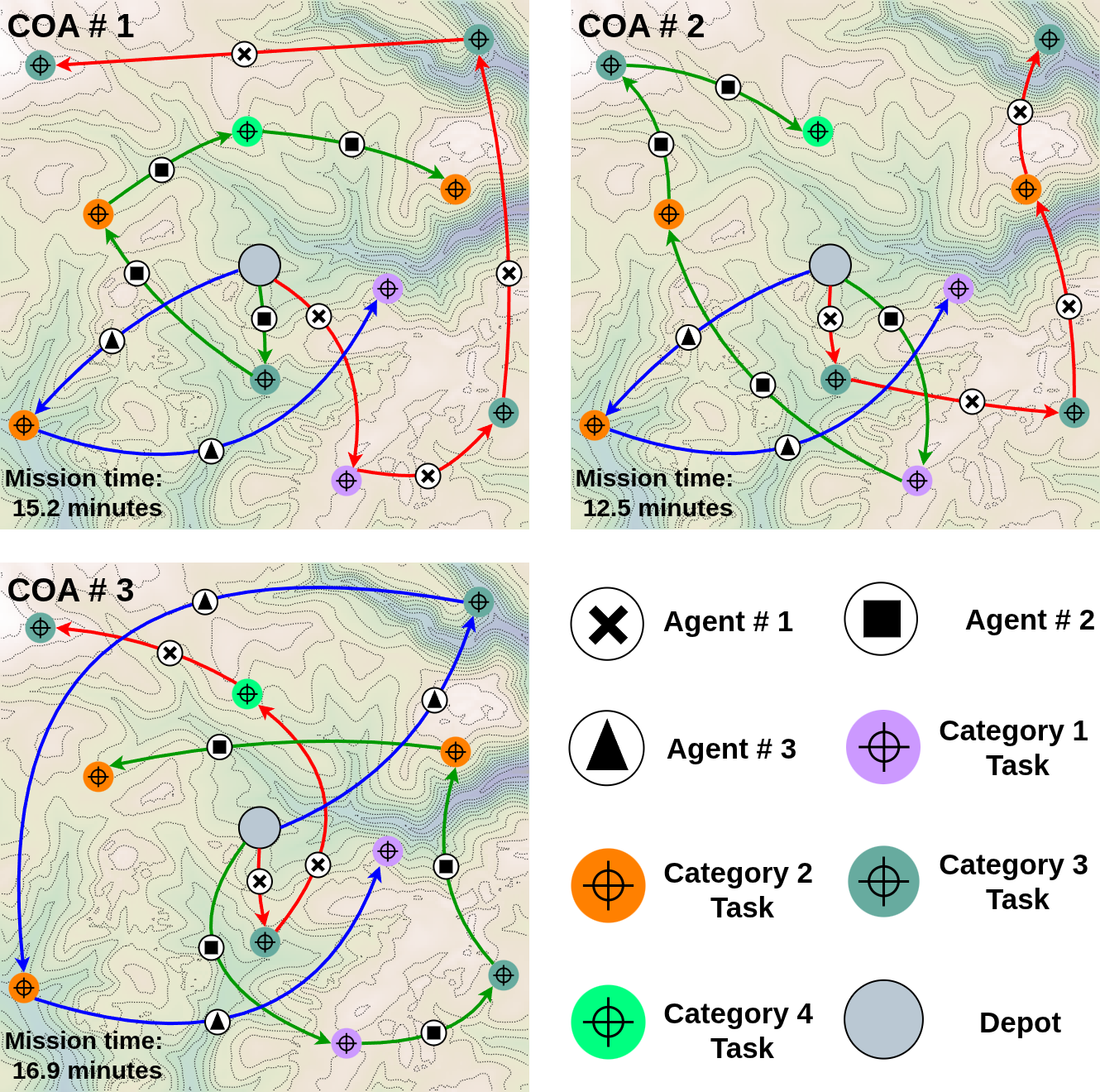}
    \footnotesize
    \caption{Example of 3 distinct candidate Courses of Action (COAs) for the same mission, involving 3 agents and 10 tasks ($\in$ 4 categories). In each COA, agents depart from a central depot and execute different task sequences.  Tasks are marked with solid circles, with color representing category. Each agent is marked by a hollow circle with a distinct symbol, over the edges expressing the task sequence (edge shape and background map are illustrative).}
    \label{fig:intro}
\end{figure}

Multiple mobile agents working together, termed as mobile collaborative multi-agent systems, promise crucial roles in solving numerous real-world problems such as disaster response \cite{disaster_response}, environmental monitoring \cite{env_monitoring}, last-mile delivery \cite{Last_mile}, Urban Air Mobility (UAM) \cite{UAM_aviation}, agricultural automation \cite{agriculture} and military operations \cite{CIL20101331}. In such complex operations, a Course of Action (COA), a concept predominantly used in military planning, serves as a predefined plan that outlines how the agents, whether humans, piloted vehicles, or autonomous vehicles/robots, should schedule and execute tasks to achieve the mission objectives. However, as the complexity of the operational environment increases due to a large number of agents, dynamic task spaces, and evolving operational constraints, automated algorithmic planning approaches are needed to support the planning of COAs. Moreover, due to the possibility of environmental changes over time, especially between when planning is done and operations are executed, it is desirable to have multiple COAs at hand as alternate options. Each of these COAs must provide a threshold level of mission performance while being distinct from each other in terms of resource/task distributions and sequences. A representative example of 3 diverse COAs for a 10-tasks/3-agents mission is shown in Figure \ref{fig:intro}. Moreover, COA planning and re-planning often needs to happen as quickly as possible prior to the mission (as opposed to a long time ahead) to be able to maximally exploit the latest information. The need to plan not one but a pool of diverse COAs thus makes it further imperative to have an automated planning method at our disposal. In this paper, we present a theoretical problem formulation and an efficient computational framework to generate such a pool of diverse COAs over a task space that presents location, deadline, and task/agent compatibility features. 

To develop such an automated approach for generating a pool of COAs, it is essential to first understand the underlying problem of planning a single COA. Given the large existing body of work in Multi-Robot Task Allocation (MRTA) problems \cite{capam,capam_tda,toptw,team_orienteering} or, more generally multi-agent task allocation (MATA), we conceive that the pre-mission process of planning or generating a single COA can be modeled as a centralized MRTA problem. In MRTA, a set of tasks must be optimally assigned to multiple robots while accounting for constraints such as time deadlines, robot capabilities, and environmental limitations \cite{iTax}, and there exists a wide spectrum of solutions ranging from heuristics \cite{team_orienteering,Mitiche2019IteratedLS, 10711564} and optimization-based techniques \cite{toptw, KIM20133065, doi:10.1287/trsc.1110.0377} to learning-based techniques \cite{capam, capam_tda, 8279546}.

Despite the significant advances in both solution domains, the need to generate multiple diverse COAs adds complexities beyond traditional MRTA formulations. Majority of the existing works focus on diversity within the behavioral space of the agents in a multi-agent setting \cite{9346741, doi.org10.1002int.23075}, while using end-to-end learning methods is not a viable option as they focus on converging upon a single optimal solution. Generating a pool of COAs invariably calls for a high computational burden that can only be partly addressed via parallelization. Moreover, very few existing approaches consider soft-heterogeneity, namely a variation in how compatible agents or agent types are to each task or task type. %While there is work on using mixture of experts to generate multiple policies for a given job, almost none of these provision the explicit capability to induce diversity across the policies.

To address these challenges and promote computational efficiency, we propose to decompose the problem of generating a pool of diverse COAs into two sub-problems: \textbf{1)} determining task assignments among agents in each COA so as to maximize diversity of pool of COAs; and \textbf{2)} optimizing the task execution sequence for each agent within each COA. We premise that diversity across COAs is driven by the differences in the resource distribution between COAs, while optimally sequencing the allocated task executions enhances the completion feasibility of each COA. 

More specifically, our proposed framework, \algo, uses a genetic algorithm (GA) that maximizes diversity among task assignments for sub-problem (1). To keep the cost of this GA-based optimization process low, we hypothesize that it must be agnostic to the operational environment, thereby eliminating the need to use a simulation to evaluate the diversity in a candidate pool of COAs. This necessitates the formulation of a new \textit{theoretical measure of diversity}. Since solely maximizing diversity risks highly skewed agent-task allocations, some of which might be infeasible to execute within given physical mission timelines, we propose to balance diversity with a measure of "\textit{compatibility}" of the agent task allocations across COAs in a candidate pool. Next, to solve sub-problem (2), \algo introduces a graph reinforcement learning method that abstracts the allocated task space for each agent in each COA as a graph and determines the sequence of tasks execution so as to maximize the overall task completion rate; this approach is conceived to readily generalize over various task features and scale over task numbers, motivated by prior work in RL-based MRTA \cite{capam,capam_tda,cam_mrta}. For the current formulation and case studies, we make the following assumptions: i) operations occur over a 2D environment with constant (nominal speed of travel for agents); ii) the multi-agent operation is of the type, Single-task Robots, Single-robot Tasks, and Time-extended Assignment (SR-ST-TA) according to \cite{iTax}, where each task has a completion deadline; iii) each agent has varying compatibilities with tasks, introducing soft heterogeneity \cite{ranjbar2013toward}, that also affect time taken to complete the task; and iv) environmental evolution used to test benefits of a COA pool is represented simply by noisy increase in travel times or random removal of connectivity between task locations. 

\textbf{Contributions:} The key contributions of this paper can thus be summarized as: \textbf{1)} Decomposition of the COA pool generation problem into two sequential sub-problems: diverse task allocations across COAs and then task sequencing in each COA; \textbf{2)} Formulating and solving the first sub-problem as a simulation-free binary optimization (with GA) that evaluates COA diversity theoretically by abstracting the COA pool itself as a graph; \textbf{3)} Posing the single agent task sequencing problem as a Markov Decision Problem, with actions provided by a graph neural net trained via policy gradient RL; and \textbf{4)} Demonstrating the effectiveness of our approach both in comparison to a random-walk baseline and optimization based task sequencing method, and in terms of sensitivity of the COAs to traverse uncertainties.

\section{PROBLEM FORMULATION}

In this paper, we address MRTA problems classified as Single-task Robots, Single-robot Tasks, and Time-extended Assignment (SR-ST-TA) \cite{iTax} (for generality, we will refer to robots as agents from here on). The goal of our approach is to generate a diverse pool of high-quality Courses of Action (COAs) for task allocation, where each COA provides a feasible assignment of tasks to agents and an optimized execution order. The two key metrics we aim to optimize are: (1) maximizing the task completion rate in each COA and (2) maximizing diversity among the COAs within the pool. The task completion rate for each COA is defined as the ratio of the number of tasks completed by all agents to the total number of tasks available in the environment. We consider each task to have an associated deadline before which it must be completed, and every agent has a fixed level of compatibility with every task that governs how long it takes for it to complete the task. Evaluating diversity, on the other hand, is not a trivial task because the COAs must be diverse in the space of how tasks are allocated to the agents, which is a combinatorial space that is not amenable to straightforward calculations of diversity in terms of direct Euclidean metrics. To this end, we develop a diversity metric that captures the spread of COAs over the combinatorial space of task allocations (mathematical details for the same have been presented in Sec.\ref{sec:GA_method}. The remainder of this section provides the necessary mathematical notations required to formulate the problem of generating a diverse pool of COAs.

\subsection{The Task Graph}
\noindent We abstract the task space and model it as a fully-connected graph $\mathcal{G}_T(t, a_j)=(N(t, a_j),E)$ where $N(t, a_j)$ represents the set of nodes/tasks and $E$ represents the set of edges connecting each every pair of tasks in $\mathcal{G}_T(t, a_j)$. The task graph, along with the task assignments for each agent in different COAs, is used to generate an optimal task-execution sequence for each agent that maximizes the number of tasks completed before their deadlines. Since the task execution order depends on the time $t$ and the decision-making agent $a_j$, $\mathcal{G}_T(t, a_j)$ is modeled as a function of both variables for all agents $j\in[1,\cdots,n_{ag}]$, where $n_{ag}$ is the total number of agents. We represent the total number of tasks as $n_{tasks}$.

In this paper, we consider each node $i$ in $\mathcal{G}_T(t, a_j)$ to be represented by a 5-dimensional feature vector $\delta_i(t, a_j)=[x_i, y_i, \nabla_i^{arr}(t, a_j), \nabla_i^{comp}(a_j),d_i(t)]$, where $(x_i,y_i)$ are the X and Y coordinates of the tasks, $\nabla_i^{arr}(t, a_j)$ is the time required by agent $a_j$ to reach task $i$ at time $t$, $\nabla_i^{comp}(a_j)$ is the time required by the agent of complete the task, and $d_i(t)$ is the time left before the task expires. The edges represent the time it would take for agent $j$ to travel between a pair of tasks, based on its velocity and the Euclidean distance between the tasks. The information abstracted within $\mathcal{G}_T(t, a_j)$ closely resembles a disaster response or a similar scenario where the nodes can be thought of as affected areas and the agents are analogous to robots delivering relief packages to the affected areas. The feature vector $\delta_i$ and the edge weights can be further extended to include additional environmental information based on the problem context.

\subsection{Defining a Course of Action}

A Course of Action (COA), denoted as $C$, is a solution to the MRTA problem that satisfies certain operational constraints. Given the task assignments for each agent, a COA represents the task execution trajectories followed by each agent and, as such, can be defined as 
% \begin{equation}
    
% \end{equation}
% where 
\begin{align}
    C &= [\tau_1,\cdots,\tau_{n_{ag}}] \text{ , where}\\
    \tau_j&=(\mathcal{T}_1,\cdots,\mathcal{T}_{|\tau_j|}) \\
    \label{eq:3}
    \text{such that }|\tau_j|&=\max_{\tau\subseteq \Gamma_j(t, a_j)}|\tau| \text{  and  } \mathbb{I}_{\mathcal{T}_i}= 1, \forall \mathcal{T}_i\in \tau_j 
    % \label{eq:4}
    % \mathbb{I}_{\mathcal{T}_i}&= 1, \forall \mathcal{T}_i\in \tau_j
\end{align}

\noindent Here, $\mathcal{T}_i\in \Gamma_j(t, a_j)$ is a task belonging to the subset of tasks $\Gamma_j(t, a_j)\subset N(t, a_j)$ assigned to agent $j$, $|\cdot|$ is the cardinality of a set, and $\mathbb{I}_{\mathcal{T}_i}$ is an indicator variable that is equal to $1$ is task $\mathcal{T}_i$ is completed before its deadline and is equal to $0$ otherwise. Eq.\ref{eq:3} ensures that the agent completes the maximum possible number of tasks assigned to it before they expire.

Using this definition, \algo focuses on automating the planning of a diverse pool of $n_{coa}$ COAs $\mathcal{C}=[C_1,\cdots,C_{n_{coa}}]$ where an elitist genetic algorithm first computes the task assignments for each agent in each COA $C_i$ and a modified version of our previously presented graph-attention based deep reinforcement learning agent, CapAM \cite{capam}, optimizes the execution sequence of assigned tasks within each $C_i$.

% $C = [\tau_1,\cdots,\tau_{n_{ag}}]$ where, $\tau_j\in \bigcup_{k=1}^{|N(t, a_j)|}\mathcal{P}_k(N(t, a_j))$, is the ordered set of tasks completed by agent $j$ such that $\bigcap_{j=1}^{n_{ag}}\tau_j=\phi$. Here, $\mathcal{P}_k(N(t, a_j))$ is the set of all permutations of length $k$ from $N(t, a_j)$, and $\bigcup,\bigcap$ represent the set union and set intersection operators respectively.

\section{METHODOLOGY}

\begin{figure*}[!t] %!t
    \centering
    \includegraphics[width=1.0\textwidth]{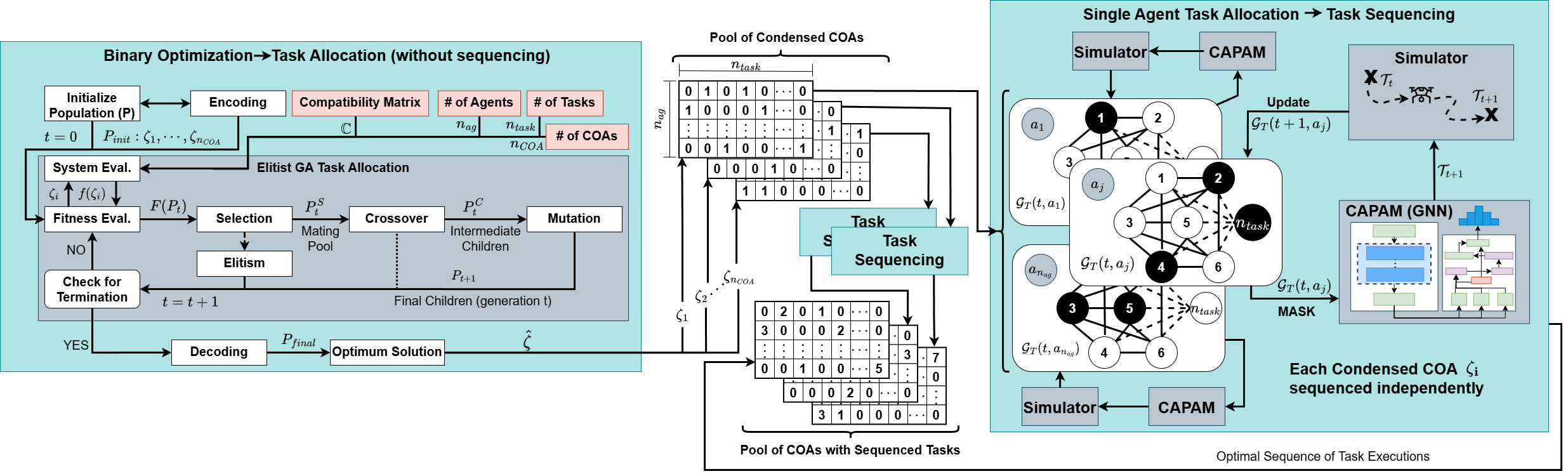}
    \footnotesize
    \caption{Complete pipeline of GenCOA. The output of the task sequencing policy marked as the pool of COAs with sequenced tasks is for illustrating the sequence in which each allotted task must be completed by the corresponding agent in each COA and does not represent the actual form of the graph-learning network's output.}
    \label{fig:architecture}
\end{figure*}

For a problem with $n_{task}$ tasks and $n_{ag}$ agents, we decompose the problem of generating a pool of $n_{coa}$ diverse COAs into two sub-problems.
\begin{itemize}
    \item Assigning tasks to each agent in each COA such that it promotes diversity within the pool. 
    \item Finding the optimal sequence in which the agents must complete the assigned tasks.
\end{itemize}

The COA generation process begins with an elitist genetic algorithm, which optimizes task assignments for each agent within every COA in the pool. Once task assignments are established, a task-sequencing network $\pi_{\theta}$  determines the optimal execution order for each agent’s assigned tasks in each COA. This sequencing process operates in parallel across all agents and all COAs. For each agent within each COA, $\pi_\theta$ iteratively selects the next task in the sequence based on the current state of the task graph. A low-fidelity simulator then simulates the execution of the selected task, updating the task graph and agent states accordingly before proceeding to the next decision step. Tasks that have already been sequenced or are predicted to expire before execution are masked from $\pi_\theta$, ensuring efficient decision-making. Figure \ref{fig:architecture} represents the entire pipeline of GenCOA, and the following subsections present additional details on the genetic algorithm and the task-sequencing network.

\subsection{Elitist Genetic Algorithm for Sub-Problem I}
\label{sec:GA_method}

Let $\hat{\zeta}=[\zeta^{(1)},\cdots,\zeta^{(n_{coa})}]$ denote a pool of $n_{coa}$ condensed COAs. A condensed COA is a mathematically abstract way of defining the task assignments for each agent and is defined as $\zeta^{(k)}\in \mathbb{B}^{n_{ag}\times n_{task}}, \zeta^{(k)}_{ij}=1$ if task $j$ is assigned to agent $i$ in COA $C_k$, and $\zeta^{(k)}_{ij}=0$ otherwise ($\mathbb{B}$ denotes the binary number system). Additionally, we define a task-agent compatibility matrix $\mathbb{C}\in \mathbb{R}^{n_{ag}\times n_{task}}$ where $0\leq\mathbb{C}_{ij}\leq 1$ is the compatibility of agent $i$ with task $j$. A higher compatibility value leads to faster task completion times and vice versa. While the compatibility matrix can be learnt from historical data, it requires non-trivial data mining and interpretation work specific to the real-world application; hence, in this paper, we assume it to be known ahead of time based on prior data or expert knowledge. In our case studies, the compatibility matrix is a prescribed entity.

Next, we pose the generation of the pool of condensed COAs as an optimization problem:
\begin{align}
    \label{eq:obj}
    \text{maximize } f(\hat{\zeta}) &= f_{\text{div}}(\hat{\zeta}) + f_{\text{comp}}(\hat{\zeta}) \\
    \label{eq:con1}
    \text{such that } \sum_{i=1}^{n_{ag}}\zeta^{(k)}_{ij}&=1, \forall j=[1,\cdots,n_{task}],\forall \zeta^{(k)}\in\hat{\zeta} \\
    \label{eq:con2}
    \sum_{j=1}^{n_{task}}\zeta^{(k)}_{ij}&\leq T_{max}, \forall i=[1,\cdots,n_{ag}],\forall \zeta^{(k)}\in\hat{\zeta}
\end{align}
where $T_{max}$ (set by the user) is the maximum number of tasks an agent is allowed to do. The objective function for the optimization problem has two components: a diversity function $f_{\text{div}}(\hat{\zeta})$ that measures the diversity and a compatibility function $f_{\text{comp}}(\hat{\zeta})$ that measures overall task-agent compatibility within the pool. 

The diversity functions inspired by \cite{9917541} and is defined as $f_{\text{div}}(\hat{\zeta})=MST(\mathcal{G}_{\hat{\zeta}})$, where $MST(\cdot)$ computes the total weight of the minimum spanning tree of a graph. $\mathcal{G}_{\hat{\zeta}}$ is a meta-graph defined as $\mathcal{G}_{\hat{\zeta}} = (\mathcal{V},\mathcal{E})$, where $\mathcal{V}=(\zeta^{(k)}|\zeta^{(k)}\in \hat{\zeta})$ is the set of nodes and $\mathcal{E}=\left(\|\mathcal{F}(\zeta^{(i)}-\zeta^{(j)})\|^2|\forall \zeta^{(i)},\zeta^{(j)}\in \hat{\zeta}\right)$ is the set of edges. $\|\cdot\|^2$ is the L-2 norm operator and $\mathcal{F}:\mathbb{R}^{a\times b}\rightarrow\mathbb{R}^{1\times ab}$ if a function that flattens a matrix. $div(\hat{\zeta})$ quantifies the diversity among COAs, and maximizing it promotes varied task assignments across the pool.

The compatibility function is defined as $f_{\text{comp}}(\hat{\zeta})=\sum_{k=1}^{n_{coa}}\sum_{i,j}(\zeta^{(k)}\odot\mathbb{C})_{ij},\forall i=[1,\cdots,n_{ag}],\forall j=[1,\cdots,n_{task}]$, where $\zeta^{(k)}\odot\mathbb{C}$ is the Hadamard product of the two matrices. $f_{\text{comp}}(\hat{\zeta})$ quantifies the overall task-agent compatibility of the pool, and maximizing it leads to agents being assigned to tasks that they are more compatible with. This, in turn, leads to faster completion of tasks.

We use an elitist genetic algorithm (GA) to solve the optimization problem. The population size of the GA is set to $100$ with a mutation probability of $0.1$, an elite preservation ratio of $0.01$, uniform crossover with a crossover probability $0.5$, and a parent ratio of $0.3$. The GA is run for $5000$ iterations. Figure.\ref{fig:combined_convhist} shows the convergence history of the genetic algorithm used for task allocations in a 2-agent, 100-tasks, 20-COAs problem.

\subsection{Graph-Attention Based Reinforcement Learning for Sub-Problem II}

After generating the pool of condensed COAs, the next step is determining the optimal sequences in which each agent must complete its tasks. We propose a task-sequencing network $\pi_{\theta}$ inspired by \cite{capam} with a graph capsule convolutional neural network encoder and a multi-head self-attention decoder, for solving what now essentially is a Single Agent Task Allocation (SATA) problem.  Since each task is assigned to a single agent with no overlap, the task-sequencing network can be executed in parallel for each agent across all COAs. The task-sequencing network is trained using Proximal Policy Optimization (PPO) \cite{ppo}, a deep reinforcement learning algorithm. Rest of this section describes the MDP formulation of the SATA problem and provides further details on the task-sequencing network.

\subsubsection{MDP representation of SATA}

The Markov Decision Process (MDP) can be defined as a tuple $\langle\mathcal{S},\mathcal{A},\mathcal{P}_a,\mathcal{R}\rangle$.

\noindent \textbf{State Space} ($\mathcal{S}$): At every decision-making instance $t$, agent $j$ has access to the task graph $\mathcal{G}_T(t, a_j)$ along with a masking vector $M(t, a_j)\in \mathbb{B}^{n_{task}\times 1}$. A decision-making instance is defined as the time $t$ when the simulator has finished simulating the latest task assignment and $\pi_\theta$ is ready to decide the next task in the sequence. The masking vector is used to hide tasks that are not available to the agent by setting $M(t, a_j)_i=0$ if task $i$ has expired/been completed / not been assigned to agent $j$, and $M(t, a_j)_i=1$ if task $i$ is available for agent $j$. This masking vector is used in the downstream processing by the task-sequencing network.

\noindent \textbf{Action Space} ($\mathcal{A}$): Is the set of actions that an agent can take, and is defined by a list of indices ${1,2,\cdots,n_{task}}$ the represent the task that the agent chooses to go to next. In scenarios involving a depot, it can be denoted by the index $0$.

\noindent \textbf{Transition Dynamics} ($\mathcal{P}_a$): The transition is an event-based trigger, and an event is defined as the condition that a new task is ready to be selected.

\noindent \textbf{Reward} ($\mathcal{R}$): We choose a reward function that penalizes the agent for every task that expires, with the penalty growing exponentially with the number of expired tasks. This incentivizes the agents to prioritize task completion before the first expiration occurs to avoid severe performance degradation. Additionally, in real-world applications, such as disaster response, missing one or two deadlines may be manageable, but as failures accumulate, system performance degrades rapidly. The exponential penalty models this behavior, ensuring the agent learns not just to minimize expired tasks but to avoid a cascade of failures. Thus we formulate the reward function $\mathcal{R}(t, a_j)$ for agent $j$ at time $t$ as:

\begin{equation}
    \mathcal{R}(t, a_j) = -e^{0.1n_{exp}(t,a_j)} + 1
\end{equation}

\noindent where $n_{exp}(t,a_j)$ is the number of tasks (out of the assigned tasks) that have expired by time $t$ for agent $j$. This reward function penalizes the age

\subsubsection{The Task-Sequencing Network (Encoder)}

\begin{figure}[h] %!t
    \centering
    \includegraphics[width=\columnwidth]{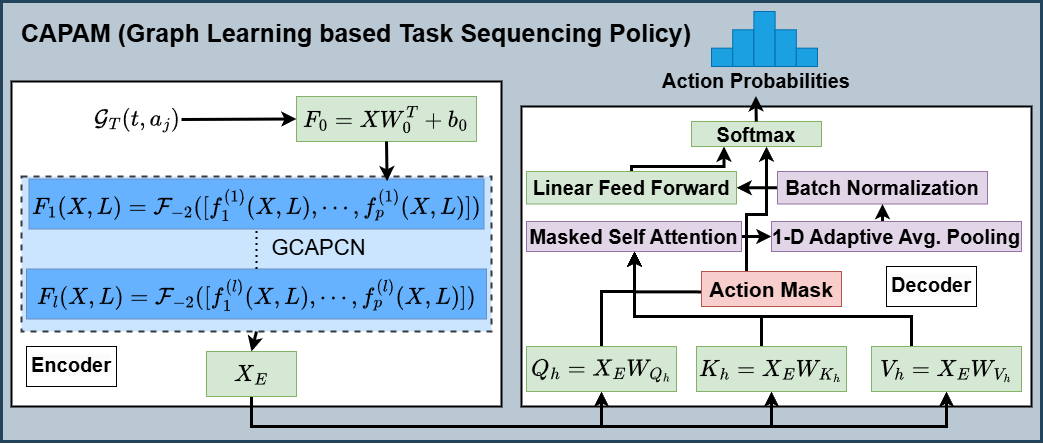}
    \footnotesize
    \caption{The encoder-decoder policy architecture by Capsule Attention Mechanism (CapAM) for single agent task sequence}
    \label{fig:capam}
\end{figure}

The task-sequencing network $\pi_\theta$, called CapAM, is an encoder-decoder architecture that was presented in our previous work \cite{capam}, and that takes the task-graph $\mathcal{G}_T(t, a_j)$ and the masking vector $M(t, a_j)$ as the input and outputs the task index that must be executed next in sequence by agent $j$. The encoder in CapAM uses a Graph Capsule Convolutional Neural Network (GCAPCN) introduced in \cite{gcaps} to generate higher-dimensional node embeddings for each node in $\mathcal{G}_T(t, a_j)$ by capturing both local and global information available within $\mathcal{G}_T(t, a_j)$. The decoder uses a masked multi-head self-attention network \cite{attention} to capture contextual relations among the node embeddings to compute probability values for each task. The task with the highest probability is chosen greedily to be the next in the task-execution sequence. Prior to the greedy selection, the masking vector is used to hide the tasks that are not available to the agent. Figure \ref{fig:capam} shows the overall architecture of the task-sequencing network, while additional details about CapAM can be found in our prior work \cite{capam}.

Given the aforementioned details, Algorithm.\ref{alg:gencoa} presents the algorithmic implementation of \algo.

\begin{footnotesize}
\begin{algorithm}
    \caption{GenCOA}
    \label{alg:gencoa}
    \footnotesize
    \begin{algorithmic} [1]
        \State \textbf{Input:} $n_{task}, n_{ag}, n_{coa}, \mathbb{C}$, Task Locations, Task Deadlines 
        \State \textbf{Output:} Pool of diverse COAs $\mathcal{C}=[C_1,\cdots,C_{n_{coa}}]$
        \State \textbf{Initialize:} Task Sequencing Network $\pi_\theta$
        % \State Solve the optimization problem in Sec.\ref{sec:GA_method}
        \State \textbf{Generate:} Pool of condensed COAs $\hat{\zeta}=[\zeta^{(1)},\cdots,\zeta^{(n_{coa})}]$ using Genetic Algorithm
        \For{each $\zeta^{(i)}$ in $\hat{\zeta}$}
            \For{each agent $a_j$'s task allocation $\zeta^{(i)}_j\in\mathbb{R}^{1\times n_{task}}$}
             \State Mask unallocated tasks
             \State \textbf{Initialize:} Simulator at initial state
             \Repeat 
             \State Compute task graph $\mathcal{G}_T(t,a_j)$ from simulator
             \State Identify next task $\mathcal{T}_{t+1}$ using $\pi_\theta$
             \State Simulate the execution of $\mathcal{T}_{t+1}$
             \State Mask $\mathcal{T}_{t+1}$ and tasks that have expired
             \Until{all tasks have been masked}
             \State $\tau_j=(\mathcal{T}_{1},\cdots,\mathcal{T}_{|\tau_j|})$
            \EndFor
            \State $C_i=[\tau_1,\cdots,\tau_{n_{ag}}]$
        \EndFor
        \State \textbf{return} $\mathcal{C}=[C_1,\cdots,C_{n_{coa}}]$
    \end{algorithmic}
\end{algorithm}
\end{footnotesize}

\subsubsection{Training the Task-Sequencing Network}

The task-sequencing network is trained using Proximal Policy Optimization (PPO). In each training iteration, for the purposes of experimental evaluation, the task locations remain fixed over an area of $1\times1$ sq. Km. This assumption reflects most of our targeted problems, like disaster response, search and rescue, military applications, etc., where the users have prior knowledge of task locations, unlike other environmental factors that may vary over time. The network is trained to achieve generalizability over different types of tasks and agents by randomly changing the task-agent compatibility and the task deadlines in each training iteration. The task compatibilities are drawn from a uniform distribution $\mathbb{C}\sim\mathcal{U}(0,1)$ and the time taken by the agent to complete a task $i$ is computed as $\nabla_i^{comp}(a_j)=10/\mathbb{C}_{ij}$, while the task deadlines are sampled from a uniform distribution $\mathbb{C}\sim\mathcal{U}(500,5\times10^4)$ seconds, for each iteration. The task compatibilities and deadlines are set heuristically and can be modified to mimic any real-world scenario \cite{capam}.

The training and evaluations are run on a workstation operating on Ubuntu 22.04 with an Intel Core i9 processor and an Nvidia RTX 3080 Ti GPU. All codes have been written in Python, and we utilize the PPO implementation under Stable Baselines3 \cite{stable-baselines3} for training the sequencing policy and PyTorch to define all neural networks. 

% \begin{figure}[h] %!t
%     \centering
%     \includegraphics[width=0.9\columnwidth]{paper_images/rewards.pdf}
%     % \setlength{\belowcaptionskip}{-10pt}
%     \caption{Convergence plot of the rewards while training the task-sequencing policy.}
%     \label{fig:training_reward}
% \end{figure}

\begin{figure}[h] %!t
    \centering
    \begin{subfigure}[b]{0.8\columnwidth}
        \centering
        \includegraphics[width=\textwidth]{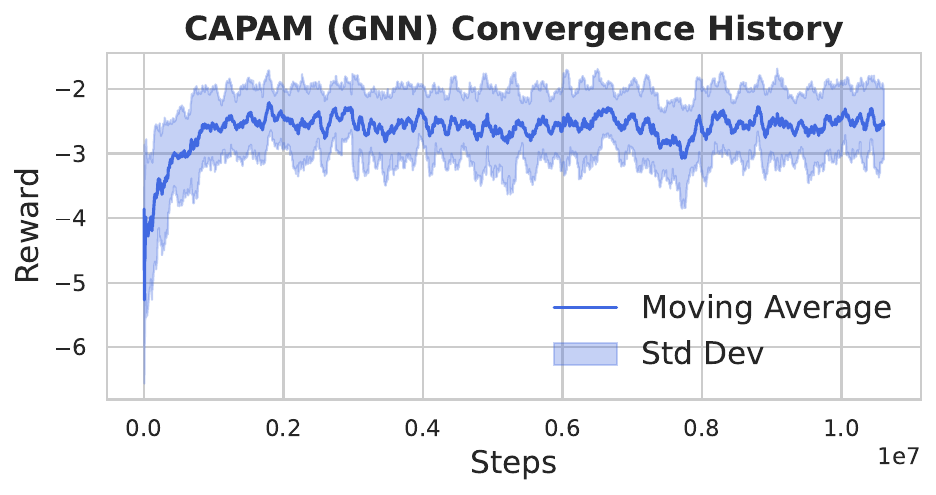}
        \caption{Training history of rewards for CAPAM training (sub-problem II).}
        \label{fig:training_reward}
    \end{subfigure}
    % \begin{subfigure}[b]{0.8\columnwidth}
    %     \centering
    %     \includegraphics[width=\textwidth]{paper_images/ga_convhist.pdf}
    %     \caption{Convergence plot of objective function for GA.}
    %     \label{fig:ga_convhist}
    % \end{subfigure}
    \begin{subfigure}[b]{\columnwidth}
        \centering
        \includegraphics[width=\textwidth]{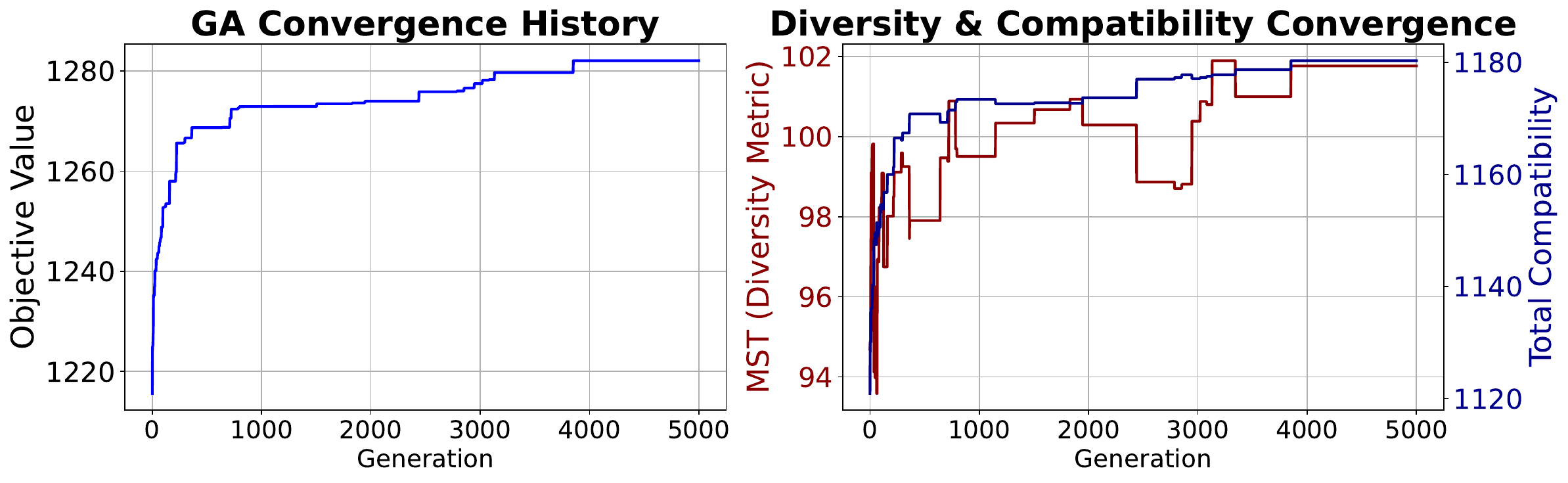}
        \caption{Training history of net objective function (left), and diversity and compatibility terms (right), for GA in the 2-agent 100-tasks 20-COAs case (sub-problem I).}
        \label{fig:combined_convhist}
    \end{subfigure}
    \footnotesize
    \caption{Training history for GenCOA: solving sub-problem I by GA (in example case) and sub-problem II by CAPAM.}
    \label{fig:training_plots}
\end{figure}

\section{Experimental Evaluation}

For our experimental evaluations, we consider an environment consisting of 100 tasks with fixed, pre-determined locations that remain unchanged over time. The task-sequencing network is trained on the entire task graph, effectively simulating a single-agent system where one agent attempts to complete all 100 tasks during training. At each training iteration, the task-agent compatibility and task deadlines are randomly assigned. Figure \ref{fig:training_reward} shows the convergence of the rewards for the training of the task-sequencing policy.

During evaluation, we introduce up to five unique agents and categorize each task into one of five task categories. The genetic algorithm allocates a subset of tasks to each agent, and each agent then employs an identical copy of the trained task-sequencing network to determine the optimal execution sequence for its assigned tasks. The task-agent compatibility is fixed based on the agent type and task category. We conduct separate experiments with 2, 3, 4, and 5 agents to analyze the impact of varying agent numbers on task completion performance. For each set of experiments, we generate a pool of 20 diverse COAs.

\subsection{Comparing With Baselines for Task Sequencing}

We begin with comparing the performance of the task sequencing network CapAM with 2 baselines: \textbf{1)} Random Walk, and \textbf{2)} Team Orienterring Problem with Time Windows (TOPTW). For this evaluation, the initial task allocation is the same for all three algorithms and is carried out by the genetic algorithm. For experiments with $n_{ag}$ agents, the maximum number of tasks that can be assigned to each agent is set to be $(100/n_{ag})+10$. Under Random Walk, each agent randomly determines the sequence of task execution. TOPTW, presented by \cite{toptw} formulates the task sequencing problem as a mixed-integer programming problem that determines the optimal sequence of tasks to maximize the number of tasks executed. We use IBM ILOG CPLEX Optimization Studio to solve the mixed integer programming problem in each experiment.

\begin{figure}[htbp]
    \centering
    \begin{subfigure}[b]{0.48\textwidth}
        \centering
        \includegraphics[width=\textwidth]{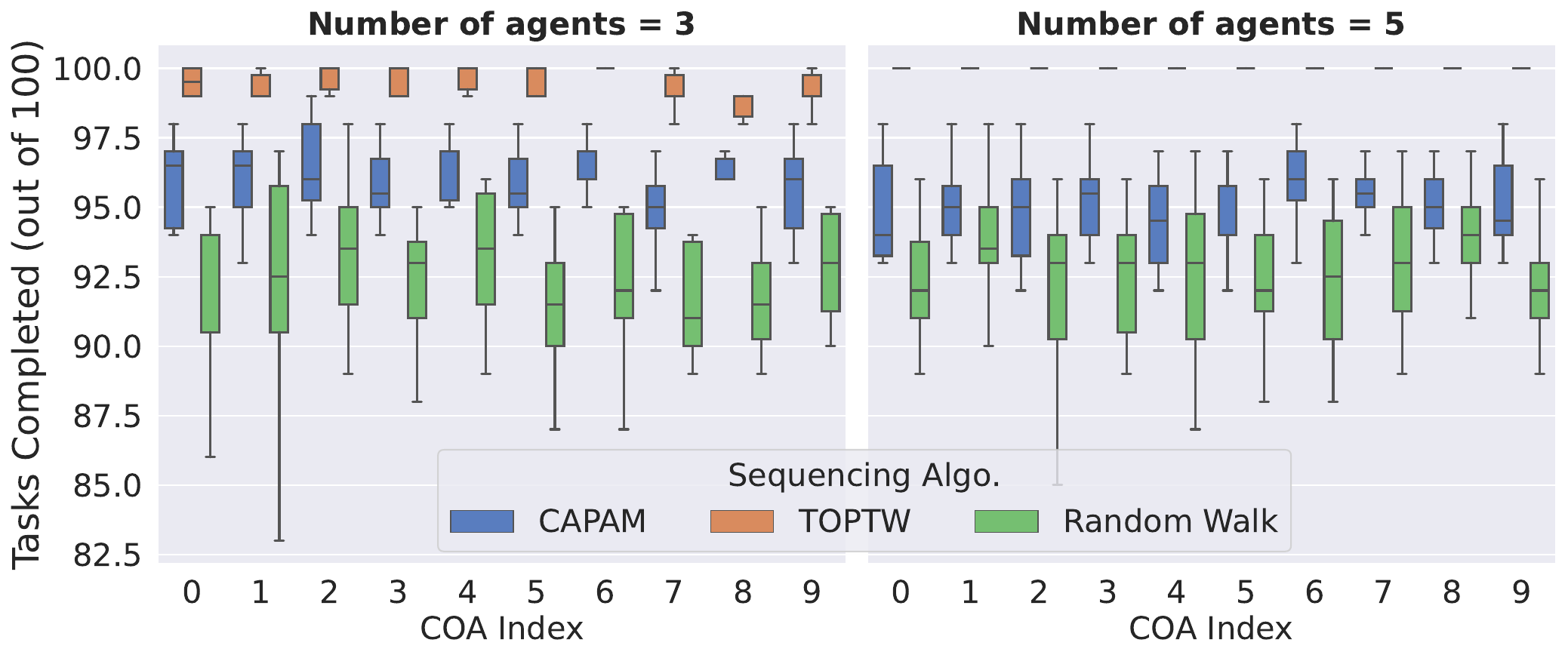}
        \footnotesize
        \caption{Performance comparison of CapAM and baselines based on tasks completed per COA. Results shown for experiments with 3 and 5 agents, plotting 10 out of 20 COAs for each experiment, for readability.}
        \label{fig:baseline_3_5_ag}
    \end{subfigure}
    
    % \vspace{0.5cm} % Adjust spacing between subfigures
    
    \begin{subfigure}[b]{0.48\textwidth}
        \centering
        \includegraphics[width=0.9\textwidth]{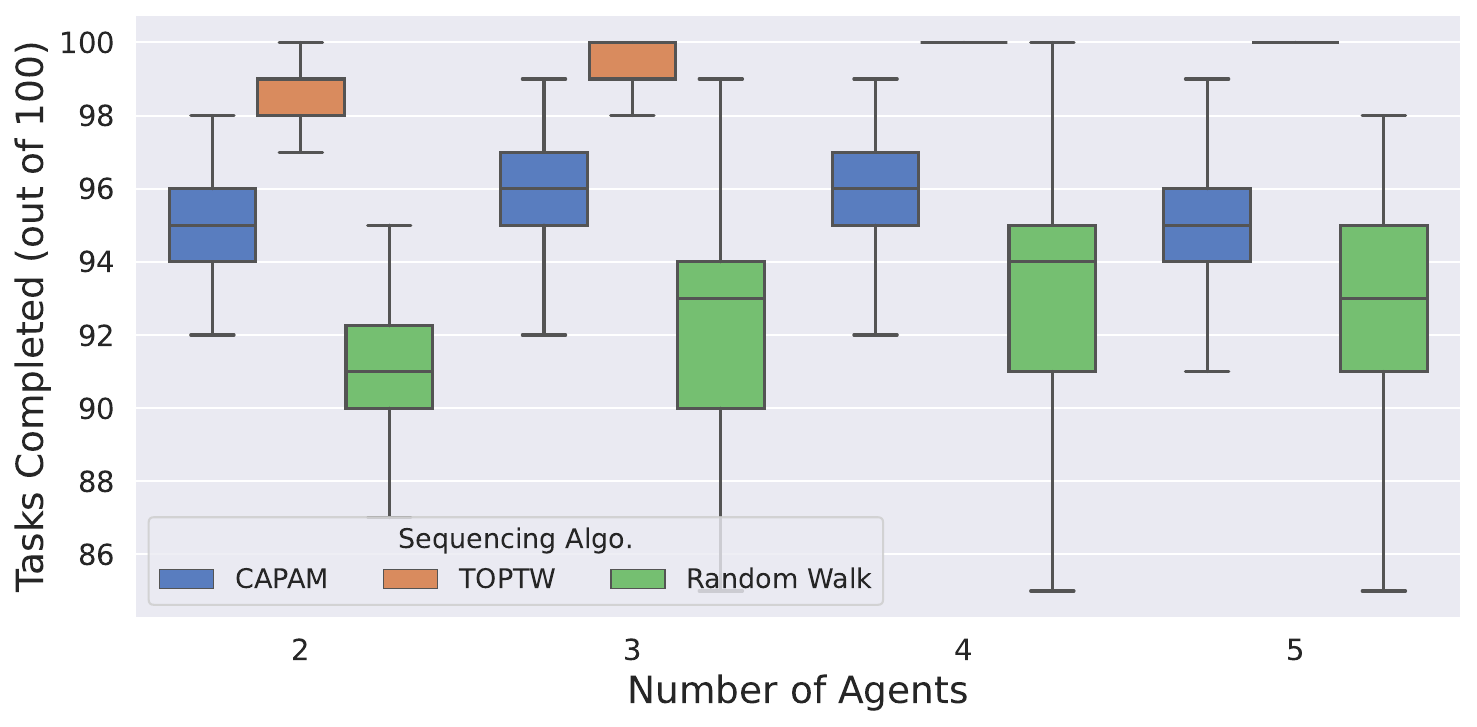}
        \footnotesize
        \caption{Comparing average tasks completed across all 20 COAs in the pool, for experiments with 2, 3, 4, and 5 agents.}
        \label{fig:baseline_3_5_ag_all}
    \end{subfigure}
    \footnotesize
    \caption{Comparison of CapAM and baselines for sequencing tasks in COAs, post task allocations performed by GA for all.}
\end{figure}

Figure \ref{fig:baseline_3_5_ag} demonstrates the number of tasks completed by each COA, under two separate experiments with 3 and 5 agents.  For each COA, results are obtained by running 10 task sequencing experiments, where task deadlines are randomly assigned in each run. From the plots, we see that CapAM performs significantly well compared to Random Walk while TOPTW performs the best in all scenarios. This aligns with expectations, as purely optimization-based approaches tend to surpass learning-based methods in such settings. Figure \ref{fig:baseline_3_5_ag_all} compares the average number of tasks completed across all COAs for experiments with 2, 3, 4, and 5 agents.  As observed previously, CapAM performs better than Random Walk, while TOPTW consistently achieves the highest performance. In scenarios with 2 and 3 agents, each agent is assigned a larger number of tasks, making it challenging even for TOPTW to complete all tasks. To quantify the deviation of CapAM from the optimal solution (TOPTW), we compute the Mean Absolute Percentage Error (MAPE). Across all COAs and all agents, CapAM achieves an average MAPE of $4.07\%$, indicating that its task completion performance is, on average, within $96\%$ of the optimal solution.

\begin{table}
\caption{Mean execution time (in seconds)  by algorithm, for varying number of agents, with standard deviation}
\label{tab:execution_time}
\begin{tabular}{|c|c|c|c|c|}
\hline
\backslashbox[1.5cm]{Algo.}{Agents} & 2 & 3 & 4 & 5 \\
\hline
CapAM & $\mathbf{0.12 \pm 0.01}$ & $\mathbf{0.09 \pm 0.01}$ & $\mathbf{0.07 \pm 0.01}$ & $\mathbf{0.06 \pm 0.01}$ \\
TOPTW & $48.07 \pm 44.72$ & $7.83 \pm 13.58$ & $2.97 \pm 9.33$ & $1.36 \pm 1.18$ \\
\hline
\end{tabular}
\end{table}

% \begin{figure}[h] %!t
%     \centering
%     \includegraphics[width=1\columnwidth]{paper_images/baseline_all_coas_times.pdf}
%     % Adjust width as needed
%     \setlength{\belowcaptionskip}{-10pt}
%     \caption{Comparing the execution times for CapAM and TOPTW.}
%     \label{fig:baseline_time}
% \end{figure}

Despite the superior performance of TOPTW in terms of the number of tasks completed, it is computationally expensive to run an optimization solver. Table \ref{tab:execution_time} compares the average execution times across all COAs in each experiment for CapAM and TOPTW. The results show that the execution time difference between the two methods can be as high as two orders of magnitude for experiments involving 2 agents and the lowest difference is an order of magnitude for experiments with 5 agents (additionally, it must be noted that each optimization run was limited to run for a maximum of 180 seconds). Additionally, the genetic algorithm takes approximately $25$ minutes to generate 20 COAs for 2 agents/100 task problems and up to $50$ minutes for 5 agents/100 task problems.

\begin{figure}[h]
    \centering
    \begin{subfigure}[b]{0.48\textwidth}
        \centering
        \includegraphics[width=\textwidth]{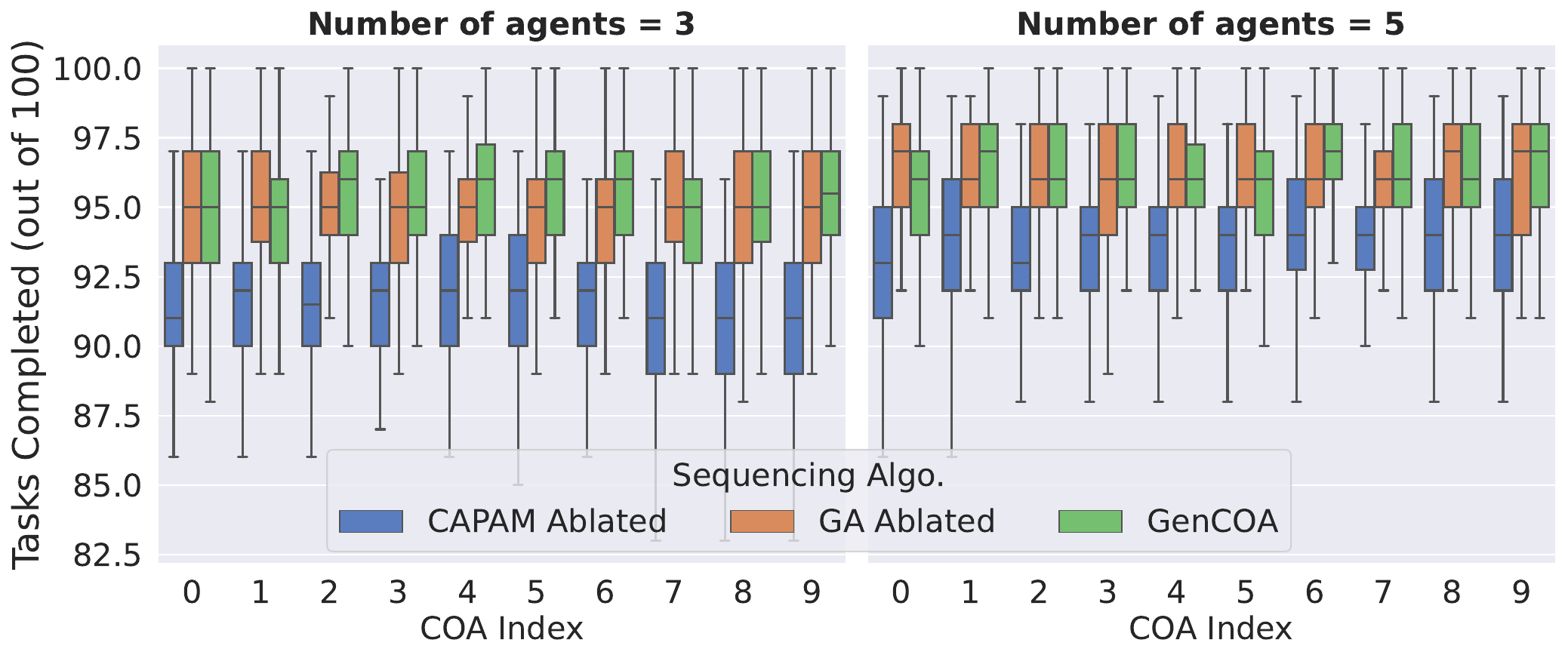}
        \footnotesize
        \caption{Performance comparison of ablated models and GenCOA based on tasks completed per COA. Results shown for experiments with 3 and 5 agents.}
        \label{fig:ablation_3_5_ag}
    \end{subfigure}
    
    % \vspace{0.5cm} % Adjust spacing between subfigures
    
    \begin{subfigure}[b]{0.48\textwidth}
        \centering
        \includegraphics[width=0.9\textwidth]{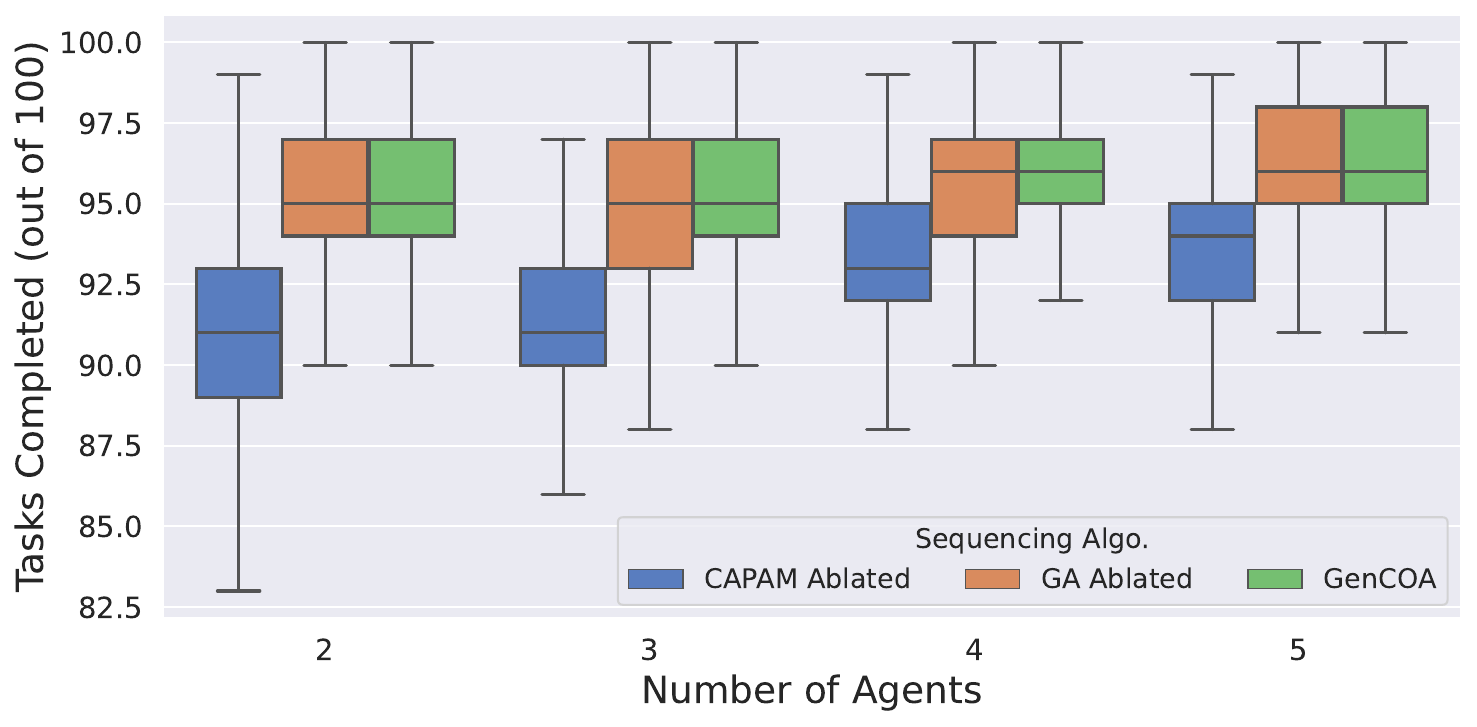}
        \footnotesize
        \caption{Comparing average tasks completed by the ablated models and GenCOA across all 20 COAs in the pool, for experiments with 2, 3, 4, and 5 agents.}
        \label{fig:ablation_all}
    \end{subfigure}
    \footnotesize
    \caption{Comparison of GenCOA and ablated models}
\end{figure}

\subsection{Ablation Study}

We perform an ablation study to evaluate the individual contributions of the genetic algorithm and CapAM in GenCOA. We compare GenCOA against two ablated models: \textbf{1)} CapAM ablated where the task sequencing network is replaced by a random walk, and \textbf{2)} GA ablated where the genetic algorithm is replaced by random task allocation. Figure \ref{fig:ablation_3_5_ag} shows the results for experiments with 3 and 5 agents. While the CapAM ablated model clearly performs the worst, the GA ablated model has a seemingly equivalent performance with GenCOA. Thus, to identify any statistically significant difference in the performance of GenCOA from the GA ablated model, we conduct pair-wise T-tests with the null hypothesis being that there is no difference in the mean total number of tasks completed by both methods. The pair-wise T-test yields a p-value $\leq 0.05$ for 14 out of 20 COAs in experiments with 3 agents and for 13 out of 20 COAs for experiments with 5 agents, suggesting that for more than $50\%$ of the COAs,  GenCOA achieves significantly better performance than the GA-ablated model.

Figure \ref{fig:ablation_all} compares the performance of all three models across all COAs for experiments with 2, 3, 4, and 5 agents. Conducting a pair wise T-test test between the GA ablated model and GenCOA yields a p-value of $2.7\times10^{-15}$ for 2 agents, $1.2\times10^{-20}$ for 3 agents, $5.6\times10^-{26}$ for 4 agents and $1.5\times10^{-15}$ for 5 agents, indicating that GenCOA has significantly better performance than GA ablated when considered across all COAs. To further analyze the contributions of GA, using the 2-agents 100-tasks 20-COAs experiment, we look at the estimated diversity within the optimized pool achieved with GA, compared to that observed in the best solutions from the initial GA populations (i.e., best of random task allocations). As also seen from Fig. \ref{fig:combined_convhist}, this improvement is found to be about 10\%, which tends to increase further if the weight of the diversity term is increased in the objective function of Sub-problem I. This substantial diversity appreciation demonstrates the utility of posing and solving Sub-problem I with GA.

Note that in this work, we use a single-objective GA for task allocation, based on an equally weighted combination of diversity and compatibility, mainly for the purpose of keeping the computing costs of solving Sub-problem I low (amenable to pre-mission planning). To further analyze how diversity and compatibility trade off with each other and potentially offer more COA options to the user, a multi-objective version of this problem can be readily solved using a standard multi-objective GA in future work, albeit with some modifications to mitigate the potential increase in computing costs.

\begin{figure}[h]
    \centering
    \includegraphics[width=\columnwidth]{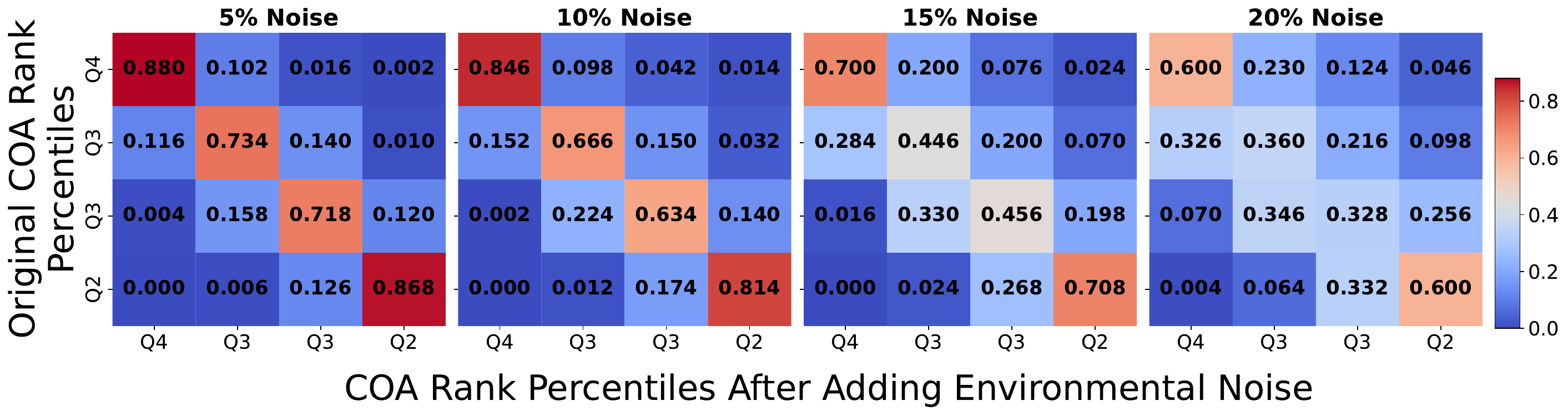} % Adjust width as needed
    \footnotesize
    \caption{Transition probability matrices for experiments with 5 agents and a pool of 20 COAs, illustrating the likelihood of a COA changing its rank after dynamic disturbances are introduced into the environment. COAs in the 4th quartile represent the top-ranked solutions, while those in the 1st quartile are the lowest-ranked.}
    \label{fig:transition}
\end{figure}

\subsection{Impact of Environmental Perturbations on COA Rankings}

 In this evaluation, we use a pool of 20 COAs for 5 agents, and lexicographically rank the COAs based on three criteria (in their decreasing order of priorities): \textbf{1)} the number of tasks completed, \textbf{2)} the total time taken, and \textbf{3)} the overall task-agent compatibility. After ranking the COAs, we categorize them into quartiles based on their rankings and analyze how frequently a COA transitions between quartiles under dynamic uncertainties. Figure \ref{fig:transition} shows the transition probabilities of COAs shifting between quartiles when noise is introduced into the environment. We track the transition probabilities for 4 experiments with $5\%, 10\%, 15\%, \text{ and }20\%$ noise levels. A $\lambda\%$ noise indicates that $\lambda\%$ of the edges of the task graph were randomly selected, and their edge weights were increased by $\lambda\%$, simulating unexpected increases in travel times for an agent along those edges. The top-ranked COAs (Q4) and bottom-ranked COAs (Q1) exhibit the highest likelihood of remaining in their respective quartiles despite the introduction of noise, suggesting that they are less sensitive to dynamic uncertainties. In contrast, mid-ranked COAs in Q2 and Q3 show a greater tendency to shift between quartiles, indicating higher sensitivity to environmental changes. Upon analyzing the performance of the best-ranked COA, we find that it stays ranked \#1 until $20\%$ noise is added, upon which it moves to rank \#2 and is replaced by a COA from the $2^{nd}$ quartile. This demonstrates how having a pool of COAs (as opposed to a single best COA) provides robustness against environmental disturbances. %A similar analysis of the COAs generated by the GA ablated model reveals a slight decrease in the likelihood that the top and bottom COAs would maintain their ranking bins when compared to GenCOA.
\begin{figure}[h]
    \centering
    \includegraphics[width=0.7\columnwidth]{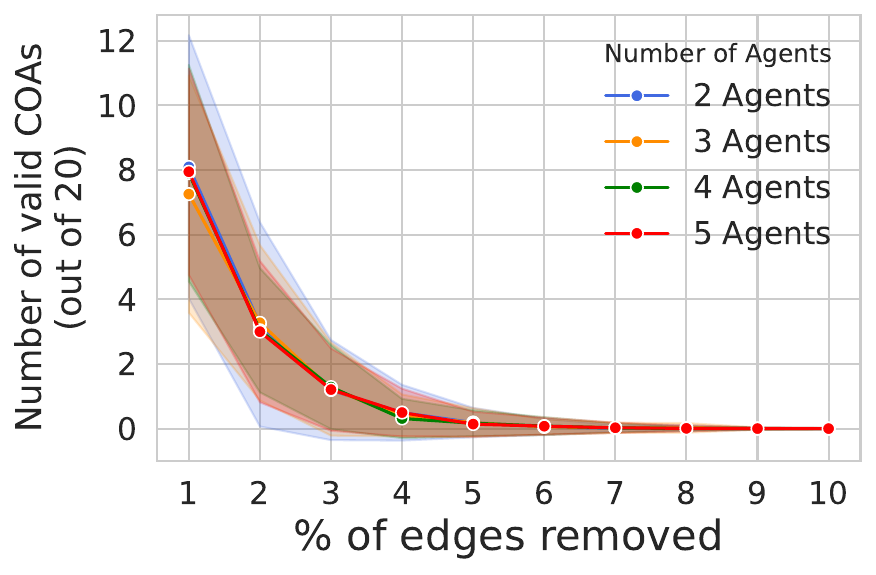} % Adjust width as needed
    \footnotesize
    \caption{Tracking the number of valid COAs out of a pool of 20 COAs when edges are randomly disabled in the task graph.}
    \label{fig:valid}
\end{figure}
% \vspace{-1cm}

Finally, we test the robustness of the pool of COAs against uncertain environmental disruptions by randomly removing edges from the task graph and tracking the number of remaining valid COAs. This simulates real-world scenarios, such as disaster response, where certain paths between tasks may become not traversable. We define that a COA becomes invalid if it requires an agent to travel across an edge that has been removed from the graph. Figure \ref{fig:valid} plots the number of valid COAs as the percentage of edges randomly removed from the task graph increases. It can be seen that for all the experiments with varying numbers of agents, at least one COA in the pool remains valid for up to $5\%$ of the edges disabled, further highlighting the benefits of having multiple alternatives as provisioned by a diverse pool of COAs.

\section{Conclusion}
%%%%%%%%%%%%%%%%%%%%%%%%%%%%%%%%%%%
In this paper, we presented a novel formulation of the problem of generating a pool of COAs for operations by multiple mobile agents, and a solution framework called GenCOA to automatically solve this problem by decomposing it into diverse task allocation and per-agent task sequencing sub-problems. Graph abstracted theoretical formulation of COA diversity and overall compatibility of the candidate allocations allowed GA based solutions to the first sub-problem within 50 mins, even for a large 20-COA pool with 5-agent/100-task operations. Subsequently, the graph RL (CapAM) trained task sequencing solutions were found to be within $96\%$ of optimal solutions given by a much slower TOPTW approach. The ablation study highlighted the significance of using a GA for task allocation, where more than $50\%$ of the COAs generated by the GA statistically outperform the COAs generated by random task allocation. Additionally, the GA boosts diversity with the pool of COAs, with the diversity in a pool of COAs for 2-agents 100-tasks 20-COAs being almost $10\%$ higher than a random task assignment. Lastly, preliminary experiments with the addition of traverse uncertainty show the importance of having multiple COAs at hand, since a subset of COAs become invalid or the COAs' ranks shift within the pool (in terms of completion performance) due to the uncertainties.  

\textbf{Future work:} Firstly, the current theoretical formulation of diversity and compatibility (at the COA pool level) are tied to the SR-ST-TA class of multi-agent task allocation problems, leaving scope for a more generalized formulation in the future, that can also tackle ``multi-robot task" and ``task-order dependency" characteristics. More detailed analysis is also needed in future work to understand and assess how diversity evolves and correlates to the robustness of the COA pool as a whole, as functions of environment-knowledge gaps modeled as uncertainties. 
%Lastly, future extension of the proposed framework to complex application-specific environments and thus expanded agent/task feature sets would help further demonstrate the benefits and needs of such automated COA pool generation approaches. %The next step would be to develop a task-sequencing methodology that is generalizable over varying task locations and can be scaled up to a large number of tasks without the need to be re-trained. Furthermore, we will work on developing a recommender system that would use the real-time physical information available from the environment as context to identify the best course of action from the pool.

\bibliographystyle{IEEEtran}

\bibliography{references}

\begin{thebibliography}{10}
\providecommand{\url}[1]{#1}
\csname url@samestyle\endcsname
\providecommand{\newblock}{\relax}
\providecommand{\bibinfo}[2]{#2}
\providecommand{\BIBentrySTDinterwordspacing}{\spaceskip=0pt\relax}
\providecommand{\BIBentryALTinterwordstretchfactor}{4}
\providecommand{\BIBentryALTinterwordspacing}{\spaceskip=\fontdimen2\font plus
\BIBentryALTinterwordstretchfactor\fontdimen3\font minus \fontdimen4\font\relax}
\providecommand{\BIBforeignlanguage}[2]{{%
\expandafter\ifx\csname l@#1\endcsname\relax
\typeout{** WARNING: IEEEtran.bst: No hyphenation pattern has been}%
\typeout{** loaded for the language `#1'. Using the pattern for}%
\typeout{** the default language instead.}%
\else
\language=\csname l@#1\endcsname
\fi
#2}}
\providecommand{\BIBdecl}{\relax}
\BIBdecl

\bibitem{disaster_response}
\BIBentryALTinterwordspacing
P.~Ghassemi and S.~Chowdhury, ``Decentralized task allocation in multi-robot systems via bipartite graph matching augmented with fuzzy clustering,'' in \emph{Proceedings of the International Design Engineering Technical Conferences and Computers and Information in Engineering Conference (IDETC-CIE)}, ser. 44th Design Automation Conference, vol.~2A, 08 2018, p. V02AT03A014. [Online]. Available: \url{https://doi.org/10.1115/DETC2018-86161}
\BIBentrySTDinterwordspacing

\bibitem{env_monitoring}
\BIBentryALTinterwordspacing
M.~V. Espina, R.~Grech, D.~De~Jager, P.~Remagnino, L.~Iocchi, L.~Marchetti, D.~Nardi, D.~Monekosso, M.~Nicolescu, and C.~King, \emph{Multi-robot Teams for Environmental Monitoring}.\hskip 1em plus 0.5em minus 0.4em\relax Berlin, Heidelberg: Springer Berlin Heidelberg, 2011, pp. 183--209. [Online]. Available: \url{https://doi.org/10.1007/978-3-642-18278-5\_8}
\BIBentrySTDinterwordspacing

\bibitem{Last_mile}
\BIBentryALTinterwordspacing
J.-P. Aurambout, K.~Gkoumas, and B.~Ciuffo, ``Last mile delivery by drones: an estimation of viable market potential and access to citizens across european cities,'' \emph{European Transport Research Review}, vol.~11, no.~1, p.~30, Jun 2019. [Online]. Available: \url{https://doi.org/10.1186/s12544-019-0368-2}
\BIBentrySTDinterwordspacing

\bibitem{UAM_aviation}
\BIBentryALTinterwordspacing
P.~Poddar, S.~Paul, and S.~Chowdhury, \emph{A Graph-Based Adversarial Imitation Learning Framework for Reliable \&amp; Realtime Fleet Scheduling in Urban Air Mobility}, 2024. [Online]. Available: \url{https://arc.aiaa.org/doi/abs/10.2514/6.2024-4006}
\BIBentrySTDinterwordspacing

\bibitem{agriculture}
\BIBentryALTinterwordspacing
H.~Guo, Z.~Miao, J.~Ji, and Q.~Pan, ``An effective collaboration evolutionary algorithm for multi-robot task allocation and scheduling in a smart farm,'' \emph{Knowledge-Based Systems}, vol. 289, p. 111474, 2024. [Online]. Available: \url{https://www.sciencedirect.com/science/article/pii/S0950705124001096}
\BIBentrySTDinterwordspacing

\bibitem{CIL20101331}
\BIBentryALTinterwordspacing
I.~Cil and M.~Mala, ``A multi-agent architecture for modelling and simulation of small military unit combat in asymmetric warfare,'' \emph{Expert Systems with Applications}, vol.~37, no.~2, pp. 1331--1343, 2010. [Online]. Available: \url{https://www.sciencedirect.com/science/article/pii/S0957417409005521}
\BIBentrySTDinterwordspacing

\bibitem{capam}
S.~Paul, P.~Ghassemi, and S.~Chowdhury, ``Learning scalable policies over graphs for multi-robot task allocation using capsule attention networks,'' in \emph{2022 International Conference on Robotics and Automation (ICRA)}, 2022, pp. 8815--8822.

\bibitem{capam_tda}
S.~Paul, W.~Li, B.~Smyth, Y.~Chen, Y.~Gel, and S.~Chowdhury, ``Efficient planning of multi-robot collective transport using graph reinforcement learning with higher order topological abstraction,'' in \emph{2023 IEEE International Conference on Robotics and Automation (ICRA)}, 2023, pp. 5779--5785.

\bibitem{toptw}
M.~Van Der~Merwe, J.~Minas, M.~Ozlen, and J.~Hearne, ``The cooperative orienteering problem with time windows,'' \emph{Optimization Online}, vol.~7, no.~11, pp. 1--11, 2014.

\bibitem{team_orienteering}
\BIBentryALTinterwordspacing
I.~Roozbeh, J.~W. Hearne, and D.~Pahlevani, ``A solution approach to the orienteering problem with time windows and synchronisation constraints,'' \emph{Heliyon}, vol.~6, no.~6, p. e04202, 2020. [Online]. Available: \url{https://www.sciencedirect.com/science/article/pii/S240584402031046X}
\BIBentrySTDinterwordspacing

\bibitem{iTax}
\BIBentryALTinterwordspacing
B.~P. Gerkey and M.~J. Matari{\'c}, ``A formal analysis and taxonomy of task allocation in multi-robot systems,'' \emph{The International Journal of Robotics Research}, vol.~23, pp. 939 -- 954, 2004. [Online]. Available: \url{https://api.semanticscholar.org/CorpusID:61299428}
\BIBentrySTDinterwordspacing

\bibitem{Mitiche2019IteratedLS}
\BIBentryALTinterwordspacing
H.~Mitiche, D.~Boughaci, and M.~L. Gini, ``Iterated local search for time-extended multi-robot task allocation with spatio-temporal and capacity constraints,'' \emph{Journal of Intelligent Systems}, vol.~28, pp. 347 -- 360, 2019. [Online]. Available: \url{https://api.semanticscholar.org/CorpusID:69802635}
\BIBentrySTDinterwordspacing

\bibitem{10711564}
E.~Zero, A.~Daniele, A.~Bozzi, S.~Graffione, and A.~E. Sindi~Morando, ``A meta-heuristic approach for capacitated vehicle routing problem in fuel distribution,'' in \emph{2024 IEEE 20th International Conference on Automation Science and Engineering (CASE)}, 2024, pp. 33--38.

\bibitem{KIM20133065}
\BIBentryALTinterwordspacing
B.-I. Kim, H.~Li, and A.~L. Johnson, ``An augmented large neighborhood search method for solving the team orienteering problem,'' \emph{Expert Systems with Applications}, vol.~40, no.~8, pp. 3065--3072, 2013. [Online]. Available: \url{https://www.sciencedirect.com/science/article/pii/S0957417412012638}
\BIBentrySTDinterwordspacing

\bibitem{doi:10.1287/trsc.1110.0377}
\BIBentryALTinterwordspacing
W.~Souffriau, P.~Vansteenwegen, G.~Vanden~Berghe, and D.~Van~Oudheusden, ``The multiconstraint team orienteering problem with multiple time windows,'' \emph{Transportation Science}, vol.~47, no.~1, pp. 53--63, 2013. [Online]. Available: \url{https://doi.org/10.1287/trsc.1110.0377}
\BIBentrySTDinterwordspacing

\bibitem{8279546}
M.~C. Gombolay, R.~J. Wilcox, and J.~A. Shah, ``Fast scheduling of robot teams performing tasks with temporospatial constraints,'' \emph{IEEE Transactions on Robotics}, vol.~34, no.~1, pp. 220--239, 2018.

\bibitem{9346741}
Y.~Yang, W.~Liu, J.~Zhou, J.~Zhou, and J.~Zhang, ``Non-cooperative-game-based multi-agent collaborative planning method for distributed generations,'' in \emph{2020 IEEE 4th Conference on Energy Internet and Energy System Integration (EI2)}, 2020, pp. 273--278.

\bibitem{doi.org10.1002int.23075}
\BIBentryALTinterwordspacing
Y.~Pan, H.~Zhang, Y.~Zeng, B.~Ma, J.~Tang, and Z.~Ming, ``Diversifying agent's behaviors in interactive decision models,'' \emph{International Journal of Intelligent Systems}, vol.~37, no.~12, pp. 12\,035--12\,056, 2022. [Online]. Available: \url{https://onlinelibrary.wiley.com/doi/abs/10.1002/int.23075}
\BIBentrySTDinterwordspacing

\bibitem{cam_mrta}
\BIBentryALTinterwordspacing
S.~Paul and S.~Chowdhury, ``Learning to allocate time-bound and dynamic tasks to multiple robots using covariant attention neural networks,'' \emph{Journal of Computing and Information Science in Engineering}, vol.~24, no.~9, p. 091005, 08 2024. [Online]. Available: \url{https://doi.org/10.1115/1.4065883}
\BIBentrySTDinterwordspacing

\bibitem{ranjbar2013toward}
B.~Ranjbar-Sahraeia, S.~Alersa, K.~Stankov{\'a}a, K.~Tuylsab, and G.~Weissa, ``Toward soft heterogeneity in robotic swarms,'' in \emph{Proceedings of the 25th Benelux Conference on Artificial Intelligence (BNAIC), Delft, The Netherlands}, 2013, pp. 7--8.

\bibitem{9917541}
A.~Behjat, N.~Maurer, S.~Chidambaran, and S.~Chowdhury, ``Adaptive neuroevolution with genetic operator control and two-way complexity variation,'' \emph{IEEE Transactions on Artificial Intelligence}, vol.~4, no.~6, pp. 1627--1641, 2023.

\bibitem{ppo}
\BIBentryALTinterwordspacing
J.~Schulman, F.~Wolski, P.~Dhariwal, A.~Radford, and O.~Klimov, ``Proximal policy optimization algorithms,'' \emph{CoRR}, vol. abs/1707.06347, 2017. [Online]. Available: \url{http://arxiv.org/abs/1707.06347}
\BIBentrySTDinterwordspacing

\bibitem{gcaps}
\BIBentryALTinterwordspacing
S.~Verma and Z.-L. Zhang, ``Graph capsule convolutional neural networks,'' 2018. [Online]. Available: \url{https://arxiv.org/abs/1805.08090}
\BIBentrySTDinterwordspacing

\bibitem{attention}
A.~Vaswani, ``Attention is all you need,'' \emph{Advances in Neural Information Processing Systems}, 2017.

\bibitem{stable-baselines3}
\BIBentryALTinterwordspacing
A.~Raffin, A.~Hill, A.~Gleave, A.~Kanervisto, M.~Ernestus, and N.~Dormann, ``Stable-baselines3: Reliable reinforcement learning implementations,'' \emph{Journal of Machine Learning Research}, vol.~22, no. 268, pp. 1--8, 2021. [Online]. Available: \url{http://jmlr.org/papers/v22/20-1364.html}
\BIBentrySTDinterwordspacing

\end{thebibliography}

% \addtolength{\textheight}{-12cm}   

\end{document}